\documentclass[11pt]{article}

\usepackage[preprint]{coling}
\usepackage{float}
\usepackage{times}
\usepackage{latexsym}
\usepackage[T1]{fontenc}
\usepackage[utf8]{inputenc}
\usepackage{microtype}
\usepackage{inconsolata}

\usepackage{graphicx}
\newcommand{\dataset}{\texttt{ParaRev}}
\title{\dataset: Building a dataset for Scientific Paragraph Revision annotated with revision instruction}

 \author{Léane Jourdan \and Nicolas Hernandez \and Richard Dufour \\
         Nantes Université, École Centrale Nantes,\\ CNRS, LS2N, UMR 6004, F-44000 Nantes, France \\
         \texttt{firstname.lastname@univ-nantes.fr}
         \AND
  Florian Boudin\\
  JFLI, CNRS, Nantes University, France\\
  \texttt{florian.boudin@univ-nantes.fr} \\\And
         Akiko Aizawa \\
  National Institute of Informatics, Japan\\
  \texttt{aizawa@nii.ac.jp} \\}

\begin{document}

\maketitle
\begin{abstract}
Revision is a crucial step in scientific writing, where authors refine their work to improve clarity, structure, and academic quality. Existing approaches to automated writing assistance often focus on sentence-level revisions, which fail to capture the broader context needed for effective modification. In this paper, we explore the impact of shifting from sentence-level to paragraph-level scope for the task of scientific text revision. The paragraph level definition of the task allows for more meaningful changes, and is guided by detailed revision instructions rather than general ones. To support this task, we introduce \dataset, the first dataset of revised scientific paragraphs with an evaluation subset manually annotated with revision instructions. 
Our experiments demonstrate that using detailed instructions significantly improves the quality of automated revisions compared to general approaches, no matter the model or the metric considered.
\end{abstract}

\section{Introduction}

In the scientific domain, writing assistance is crucial as researchers share their findings through articles published in conferences or journals. However, writing articles is challenging and time-consuming, notably for non-native English speakers or young researchers~\citep{amano2023manifold}.

The field of writing assistance has grown rapidly to address these challenges leading to the development of various tools (\href{https://www.grammarly.com/}{Grammarly}, 
\href{https://www.trinka.ai/}{Trinka AI}\footnote{\href{https://www.grammarly.com/}{https://www.grammarly.com/}, \href{https://www.trinka.ai/}{https://www.trinka.ai/}}, \ldots) and specialized workshops (\href{https://in2writing.glitch.me/}{In2Writing}, \href{https://sites.google.com/view/wraicogs1}{WRAICOGS}\footnote{\href{https://in2writing.glitch.me/}{https://in2writing.glitch.me/}, \href{https://sites.google.com/view/wraicogs1}{https://sites.google.com/view/wraicogs1}}).

\begin{figure}[t]
  \includegraphics[width=\columnwidth]{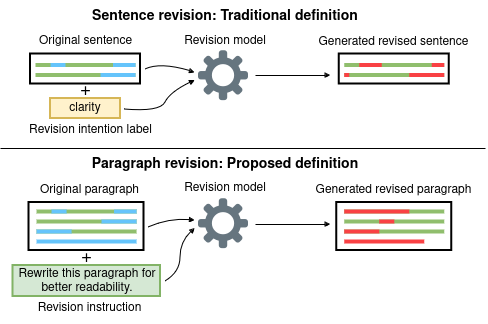}
  \caption{Definitions of the traditional sentence revision task and the proposed paragraph revision task.}
  \label{fig:shift_scope}
\end{figure}

The goal of writing assistance is to support researchers throughout the writing process, which includes four steps: Prewriting, Drafting, Revising, and Editing~\citep{jourdan2023text}. 
This paper focuses on the revision task where an input text is substantially modified for clarity, simplicity, style, and other aspects~\cite{du-etal-2022-read,li-etal-2022-text}.
 Since poor writing quality undermines the communication of research findings and often leads to paper rejection~\citep{amano2023manifold}, effective revision is a critical step in scientific writing.

Due to past limitations in processing long texts, prior research has focused on the sentence revision task (see Figure~\ref{fig:shift_scope}). In this task, a sentence is given to a seq2seq model or a Large Language Model (LLM) along with a general revision prompt, which could take the form of a label (e.g., Coherence, Style)~\citep{du-etal-2022-understanding-iterative,jiang-etal-2022-arXivEdits} or a general instruction~\cite{raheja2023coedit}. In this definition of the task, labels are assigned to specific modifications within a sentence, targeting particular spans of text to revise.

Thanks to the recent advances in NLP in the past years, we propose to expand the traditional scope of this sentence-level paradigm to detailed personalised instructions guiding the model on revisions to conduct at the paragraph level, as illustrated in Figure~\ref{fig:shift_scope}.

We argue that this new paradigm aligns better with how human writers revise the text and how LLMs are used today, allowing more comprehensive changes such as merging, splitting, or reorganizing sentences.
Additionally, personalised instructions enable more nuanced control over the degree of revision, specifying whether minor edits or major restructuring is required. They can also target specific areas within a paragraph, while other sentences provide essential context.

\begin{figure}[t]
\centering
  \includegraphics[width=0.8\columnwidth]{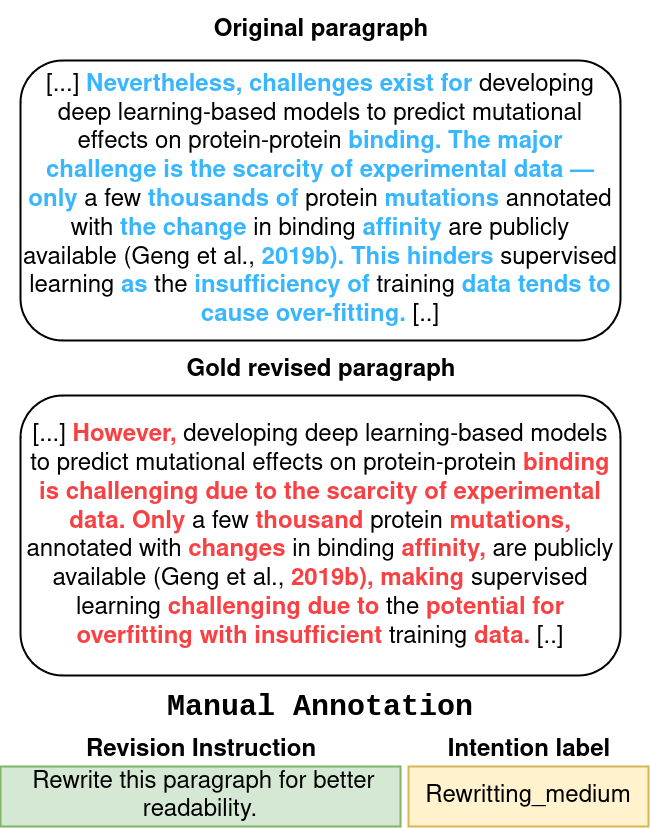}
  \caption{Example of a revised paragraph with its associated revision instruction and label.}
  \label{fig:exemple_parag}
\end{figure}

To support this task, we introduce \dataset, a corpus of paragraphs revised by their authors annotated with human revision intention labels and instructions (e.g. in Figure~\ref{fig:exemple_parag}). Our contributions are as follows:
\begin{enumerate}
    \item We proposed a definition of the text revision task at paragraph-level, with personalised revision instructions.
    \item We release a high-quality corpus of 48k revised paragraphs with an evaluation subset of 641 manually annotated paragraphs, facilitating future research in this area~\footnote{\href{https://huggingface.co/datasets/taln-ls2n/pararev}{https://huggingface.co/datasets/taln-ls2n/pararev}}.
 \end{enumerate}

\section{Related work}
Existing corpora for scientific text revision provide aligned versions of revised texts, with varying scope. Some datasets focus only on the abstract and introduction sections of scientific papers~\citep{du-etal-2022-understanding-iterative, mita-etal-2024-towards, ito-etal-2019-diamonds}, while others include full-length articles~\citep{kuznetsov-etal-2022-revise, jiang-etal-2022-arXivEdits, darcy2023aries, jourdan-etal-2024-casimir}. Most of these resources align revisions at the sentence level, though paragraph-level reconstruction is possible to capture broader, more substantial revisions.

However, not all datasets include revision annotations with explicit intention labels. Some, such as those designed for tasks related to peer-review~\citep{kuznetsov-etal-2022-revise,darcy2023aries}, focus on tracking changes without offering structured guidance for the revision process. In revision tasks, having an explicit revision intention is crucial for guiding models in performing meaningful modifications. In sentence-level revision datasets, individual modifications (i.e. spans of text) are commonly associated with a label indicating the revision intention. The taxonomies for these labels can vary across corpora~\citep{jiang-etal-2022-arXivEdits, du-etal-2022-understanding-iterative}. However, labels associated with short spans of text often lack the contextual information needed for more substantial, long-range revisions. They also do not provide the specificity that detailed instructions could offer to guide more precise edits. 

Recent efforts have attempted to bridge this gap by converting labels into general instructions to better align with how LLMs are utilized for revision~\citep{raheja2023coedit}. Our work aims to extend this approach by introducing detailed, personalized paragraph-level instructions that provide richer contextual and precise guidance for revisions.

\section{Dataset construction}

\begin{figure}[t]
  \includegraphics[width=\columnwidth]{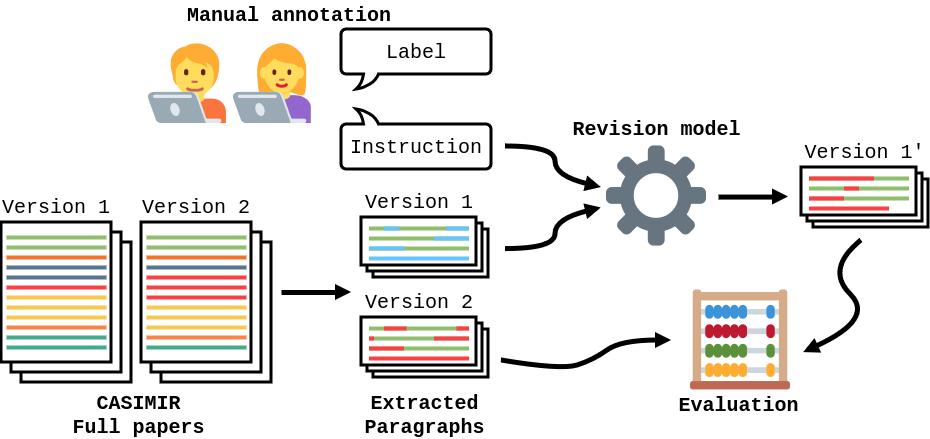}
  \caption{The data pipeline: annotation, paragraph revision and evaluation}
  \label{fig:data_pipeline}
\end{figure}

Figure~\ref{fig:data_pipeline} summarizes the overall data pipeline described in this section.

\subsection{Paragraph Selection and Extraction}
\label{subsec:selection}

Our dataset consists of pairs of revised paragraphs extracted from the CASIMIR corpus~\citep{jourdan-etal-2024-casimir}, a large resource containing revised scientific articles aligned at sentence level. This corpus provides paragraph-level IDs for each sentence, which allows us to treat paragraphs as coherent units marked by changes in paragraph IDs across both versions of the text.

However, many articles in CASIMIR contain identical or minimally revised content, which is not suitable for our purpose. We aim to build a high-quality dataset by selecting paragraphs with substantial revisions (beyond minor grammatical fixes) while preserving the original idea of the text.

To achieve this, we developed hand-crafted heuristics through empirical observations of a subset of the corpus, to retain only the sufficiently revised paragraphs (see Appendix~\ref{apx:criteria}).
From the original 1~889~810 paragraph pairs with at least one modification, we kept after this selection process 48~203 paragraphs. Extraction code is openly available~\footnote{\href{https://github.com/JourdanL/pararev}{https://github.com/JourdanL/pararev}}.

\subsection{Paragraph revision taxonomy}

To align with prior research and facilitate analysis or example selection for few-shot tasks, we chose to assign revision intention labels to each paragraph pair. Motivated by the works of \citet{du-etal-2022-understanding-iterative} and \citet{jiang-etal-2022-arXivEdits}, we propose a new paragraph-level taxonomy based on their existing sentence-level ones and observations done on a subset of our dataset.

In this taxonomy, we identified nine revision intentions, defined in Appendix~\ref{apx:taxonomy}: \textit{Rewriting (light, medium, heavy), Concision, Development, Content (addition, substitution, deletion)} and \textit{Unusable}. These labels are not associated with individual edits: they instead represent the overall revision intention for the paragraph.
Each paragraph can receive up to two labels, as multiple revisions with different intentions may occur within a single paragraph.

\subsection{Instructions}

An instruction is provided only when no new information is introduced in the revised paragraph, as revision models are only supposed to improve existing text and not make up new content. Labels are used to identify the paragraphs that do not require an annotation, i.e. the paragraphs annotated with \textit{Development}, \textit{Content Addition}, or \textit{Content Substitution}.

Annotators are asked to write concise, simple instructions as they would when guiding an LLM to revise the first version of the paragraph into the second. Detailed lists of changes are not allowed.
They must also indicate the position and intensity of revisions when necessary, especially when only part of the paragraph requires revision while the rest provides context.

Some examples of instructions and their associated pair of paragraphs are available in Appendix~\ref{apx:exampleinstruct}.

\subsection{Annotation}

The annotation process involved 10 annotators (2 professors, 3 PhD students, and 5 master's students), all not native from English and specialized in the NLP domain and experienced in reading and writing academic papers.  Most paragraphs (73.32\%) were double annotated. 

Since annotators could assign up to two labels, with 1.2 labels on average per paragraph per annotator, we used Krippendorff’s alpha for agreement. It often occurs that some revisions are on the line of two categories, e.g., \textit{Rewriting light} and \textit{medium}. Given this ambiguity, we computed two scores: one for the strict taxonomy (agreement of 0.499) and another for broader super-labels, i.e. merging similar categories (agreement of 0.693), see Appendix~\ref{apx:supercat}. Agreement with super-labels exceeds the 0.67 threshold for tentative conclusions about the consistency of the annotations~\citep{krippendorff2018content}.

Additionally, 75.32\% of paragraphs share at least one label between annotators with strict taxonomy, rising to 95.11\% using super-labels. 

Those results reflect the inherent complexity of the annotation task.

\section{Dataset Statistics}
\begin{figure*}[t]
\centering
  \includegraphics[width=\textwidth]{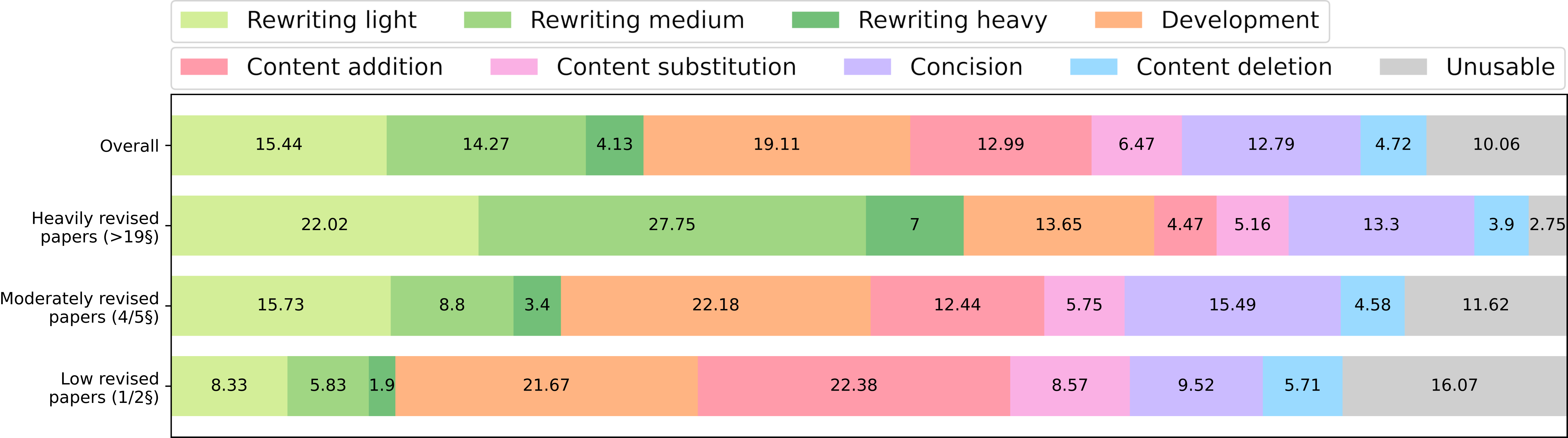}
  \caption{Distribution of labels across the dataset overall and degree of modification of the articles. }
  \label{fig:distrib_labels}
\end{figure*}

The dataset contains 48~203 paragraph pairs from 16~664 pairs of revised articles. From this total 48K paragraphs, 641 were manually annotated (470 were double annotated). This subset was chosen to represent the overall corpus based on paper revision extent:  218 paragraphs are from heavily revised papers (where over 19 paragraphs are revised), 213 from moderately revised papers (4-5 revised paragraphs), 210 from low revised papers (1-2 revised paragraphs).

Figure~\ref{fig:distrib_labels} shows the label distribution across the dataset. For fairness in the analysis, when annotators picked two labels, they were weighted 0.5 each. Additionally, paragraphs with only one annotation are counted twice.

The figure distinguishes between paragraphs from articles with different degrees of revision. Heavily revised papers tend to mainly feature \textit{Rewriting} revisions, suggesting that the entire document was evenly reworked. In contrast, low-revised papers are more likely to involve small content modifications, such as adding or removing forgotten information.

Finally, we report the instructions' distribution as follows: of the 641 annotated paragraphs, 328 have no instruction, 55 have one, and 258 have two. These 258 paragraphs form our evaluation set in Section~\ref{sec:exp}.

\section{Impact of task definition on revision}
\label{sec:exp}

\begin{table*}
  \centering
  \begin{tabular}{|c| r r | r r | r r |}
    \hline
    Metric& \multicolumn{2}{c|}{\textbf{ROUGE-L}} & \multicolumn{2}{c|}{\textbf{SARI}} & \multicolumn{2}{c|}{\textbf{Bertscore}} \\\hline
    Approach & Label & Instruction & Label & Instruction & Label & Instruction\\\hline
    CopyInput- no edits & \multicolumn{2}{c|}{78.49} & \multicolumn{2}{c|}{60.69} & \multicolumn{2}{c|}{95.98} \\
    coedit-xl & 67.50 & 67.70 & 39.56 & 39.68 & 93.88 & 93.93 \\
    Mistral-7B-Instruct-v0.2 & 45.70 & $48.23^{\dag}$ & 28.47 & $30.43^{\dag}$ & 91.38 & $91.78^{\dag}$ \\
    Meta-Llama-3-8B-Instruct & 50.37 & $55.73^{\dag}$ & 30.59 & $35.07^{\dag}$ & 91.84 & $92.68^{\dag}$ \\
    GPT4o & 57.99 & $66.17^{\dag}$ & 33.33& $41.39^{\dag}$ & 92.89 & $94.11^{\dag}$\\\hline
    Average gain &  \multicolumn{2}{c|}{+4.07}  & \multicolumn{2}{c|}{+3.66} &  \multicolumn{2}{c|}{+0.75} \\\hline
  \end{tabular}  
  \caption{Results on the paragraph revision task. Symbol ${\dag}$ marks a significative improvement.}
  \label{tab:results}
\end{table*}

To verify our hypothesis that using detailed instructions better guides the revision process compared to generic instruction labels, we conducted a comparative experiment.
For this, we evaluated how different models performed when given either a general prompt mapped from an intention label or a personalised instruction tailored to the specific changes needed (see Appendix~\ref{apx:prompt}).

We experimented with multiple models to ensure the results were robust across various architectures:
\href{https://huggingface.co/grammarly/coedit-xl}{\textbf{CoEdit}}\footnote{\href{https://huggingface.co/grammarly/coedit-xl}{https://huggingface.co/grammarly/coedit-xl}}, a T5-based model fine-tuned on sentence revision task~\citep{raheja2023coedit},
as well as \textbf{\href{https://huggingface.co/meta-llama/Meta-Llama-3-8B-Instruct}{Llama3}\footnote{\href{https://huggingface.co/meta-llama/Meta-Llama-3-8B-Instruct}{https://huggingface.co/meta-llama/Meta-Llama-3-8B-Instruct}}, \href{https://huggingface.co/mistralai/Mistral-7B-Instruct-v0.2}{Mistral}\footnote{\href{https://huggingface.co/mistralai/Mistral-7B-Instruct-v0.2}{https://huggingface.co/mistralai/Mistral-7B-Instruct-v0.2}}, and GPT-4o,} state-of-the-art foundation models with strong language understanding and generation capabilities. All models are used in zero-shot, the prompt used is given in Appendix~\ref{apx:prompt}.

Additionally, as a control baseline, we included a \textbf{CopyInput} method, which does not apply any edits to the input paragraph.

To assess the quality of revisions, we employed traditional sentence revision metrics, ROUGE-L~\citep{lin-2004-rouge} and SARI~\citep{xu-etal-2016-optimizing}, alongside Bertscore~\cite{zhang2020bertscore} to measure similarity between the generated and gold revised paragraphs. The results are summarized in Table~\ref{tab:results}.

Across all models, we observed consistent improvements when using detailed instructions over general prompts. They are even statistically significant for Mistral, Llama3, and GPT-4o, with p-values below 0.05 (paired Student's t-test).

The experiment confirms our hypothesis: instructions that provide specific revision guidance allow the models to produce more accurate revisions compared to relying solely on general labels.

However, when examining the performances of the models, we observe that the CopyInput and Co-edit achieve the best results. A manual overview of a subset of outputs reveals that Co-edit only suggests minor changes, such as grammar corrections, while other models propose more substantial modifications.

Evaluation remains a significant challenge in the text revision domain, as widely used metrics compare the proposed revision to a single reference version. This approach penalizes revisions that deviate from the gold standard, even if they result in valid improvements. Consequently, unless the model's modifications exactly replicate those made by the original author, the score will be lower than proposing no modifications (CopyInput). This limitation need to be address in future work to develop more robust and reliable evaluation methods for this task.

\section{Conclusion}

We proposed a definition of the scientific text revision task at paragraph-level, enabling more context-aware revisions using full-length instruction. Additionally, we presented \dataset, a dataset of revised paragraphs, with an evaluation split annotated with revision instructions. Our experiments demonstrate that providing detailed personalised instructions leads to more effective revisions than general ones, across multiple models.

In future work, as manual annotation is costly and time-consuming, we aim to annotate the remaining non-annotated wide split of the dataset automatically. This silver dataset will then be used to fine-tune an open-source model specifically for paragraph-level revision tasks. 

\section{Limitations}
The primary limitation of this work is the size of the evaluation subset, as it was manually annotated by volunteer researchers whose availability constrained the number of annotations. A larger annotated subset would enhance the reliability of our evaluation, allowing us to determine if smaller improvements in revision scores are statistically significant.

While the core focus of this study is on introducing personalized annotated instructions, we also labelled paragraphs with revision intention labels. 
Labelling revisions is a challenging task since multiple modifications can occur within a single paragraph, and annotators may interpret boundaries between similar categories differently. However, this limitation can be mitigated in practice by using super-labels or considering the union of the two annotations.

\section{Ethical Considerations}
\paragraph{Data availability}
All the data are extracted from the CASIMIR corpus, collected from OpenReview where all articles fall under different "non-exclusive, perpetual, and royalty-free license"~\footnote{\href{https://openreview.net/legal/terms}{https://openreview.net/legal/terms}}.

\paragraph{Computational resources}
Our experiments with revision models ran CoEdit on a local GPU for approximately two hours, while Mistral and Llama ran for nine hours on the supercomputer Jean Zay, emitting less than 0.001 tons of $CO_{2}$, with an additional 3.16\$ spent on GPT API credits.

\paragraph{Use of revision models}
We release this dataset to support future research on writing assistance for researchers. We believe that revision models based on LLMs should be used as tools to enhance clarity and structure, not to generate the primary content and analysis.
 
\section*{Acknowledgments}
We thank Jiahao Huang, Xanh Ho, Juan Junqueras,  Ken Kim, Jonas Luhrs, Julian Schnitzler and Tomás Vergara Browne for their participation in annotating the dataset.

This work was granted access to the HPC resources of IDRIS under the allocation 2023-AD011013901R1 made by GENCI.

\bibliography{custom}

\appendix

\section{Paragraph selection criteria}
\label{apx:criteria}

We keep only paragraphs that met the following requirements:
Criteria for selection (threshold obtained empirically):
\begin{itemize}
    \item \textbf{Size:} The longer version must at least be 250 characters
    \item \textbf{Percentage of modification:} 
    \begin{itemize}
        \item The most edited sentence should be at least modified at 25\%
        \item The whole paragraph should be at least edited at 10\%
        \item In a paragraph, the set of sentences modified at more than 90\% should not represent more than 40\% or 200 characters in the whole paragraph
        \item If a paragraph does not contain sentences revised at more than 50\%: The set of modified sentences should be modified at least by 20\%

    \end{itemize}
    \item \textbf{Quantity of transcribed equations:} The quantity of transcribed equations captured by regular expression should not represent more than 9\% of the set of modified sentences in the paragraph.

    \item If the paragraph starts with a modification: We check that it is not a segmentation mistake
    \begin{itemize}
        \item Is the beginning of the sentences correctly formed.
        \item If only one sentence was completely added or deleted: Accepted if it is only tags
        \item If the sentence is revised at more than 50\%
        \begin{itemize}
            \item Refused if the shorter version is equal to the end of the longer one
            \item Refused if the longer version is more than 3 times the length of the shorter one
        \end{itemize}
        \item If the sentence is revised at less than 50\%
        \begin{itemize}
            \item If the modification is at the beginning on both sides: Refused if the shorter version is equal to the end of the longer one 
            \item If the modification is at the beginning on one side: Refused if the modification is longer than 10 characters (without spaces and tags)
        \end{itemize}
        
    \end{itemize}
     \item If the paragraph ends with a modification: We check that it is not a segmentation mistake
     \begin{itemize}
        \item Is the end of the sentences correctly formed
        \item If only one sentence was completely added or deleted: Always rejected. A second version of the function exists to include cases where a full correctly formed sentence is deleted/added, resulting in 11k additional paragraphs in the corpus.
        \item If the sentence is revised at more than 50\%
        \begin{itemize}
            \item Refused if the shorter version is equal to the beginning of the longer one
            \item Refused if the longer version is more than 3 times the length of the shorter one
        \end{itemize}
        \item If the sentence is revised at less than 50\%: Always accepted
          
    \end{itemize}
    \item Check if a part of the text has not been transformed into a tag during PDF conversion 
\end{itemize}

\section{Paragraph revision taxonomy}
\label{apx:taxonomy}
See Table~\ref{tab:taxonomy}
\begin{table*}
  \centering
  \begin{tabular}{lll}
    \hline
    \textbf{Type} & & \textbf{Description} \\
    \hline
                   & Light         &   Minor changes in word choice or phrasing.\\
    Rewriting      & Medium        &  Complete rephrasing of sentences within the paragraph.  \\
                   & Heavy         &  Significant rephrasing, affecting at least half of the paragraph.\\\hline
    Concision      &               &Same idea, stated more briefly by removing unnecessary details.\\\hline
    Development    &           & Same idea, expanded with additional details or definitions.\\\hline
                   & Addition      &     Modification of content through the addition of a new idea.\\
    Content        & Substitution  &    Modification of content through the replacement of an idea or fact. \\
                   & Deletion      &   Modification of content through the deletion of an idea.  \\\hline
        Unusable  &       &  Issues due to document processing errors (e.g., segmentation problems,\\
                   &      & misaligned paragraphs, or footnotes mixed with the text). \\\hline
  \end{tabular}
  \caption{Taxonomy of revisions at paragraph level}
  \label{tab:taxonomy}
\end{table*}

\definecolor{bluesource}{HTML}{38B6FF}
\definecolor{redtarget}{HTML}{F94144}
\section{Examples of instructions}
\label{apx:exampleinstruct}
See Table~\ref{tab:examples}.
\begin{table*}
  \centering
  \begin{tabular}{p{0.25\linewidth}p{0.25\linewidth}p{0.25\linewidth}p{0.25\linewidth}}
    \hline
    \textbf{Type} &\multicolumn{3}{|l}{\textbf{Instruction}} \\
    \hline
      \multicolumn{2}{l|}{Parag source} & \multicolumn{2}{l}{Parag target}  \\\hline\hline
      \textbf{Rewriting\_light} &\multicolumn{3}{|l}{\textbf{Improve the english in the paragraph, make it slightly more formal.}} \\
    \hline
      \multicolumn{2}{p{0.5\linewidth}|}{[\ldots] Therefore, the generalization rapidly decreases after \textcolor{bluesource}{augmentationinterrupted when} training with a single background because the learning direction toward generalization about various backgrounds is not helpful to train. \textcolor{bluesource}{On the other hand,} the training can \textcolor{bluesource}{have helpwhen} their \textcolor{bluesource}{difﬁculty} is solved by \textcolor{bluesource}{augmentation, such as Figure 2(b) and Figure 2(c).} [\ldots]} & \multicolumn{2}{p{0.5\linewidth}}{[\ldots] Therefore, the generalization rapidly decreases after \textcolor{redtarget}{augmentation is interrupted during} training with a single background because the learning direction toward generalization about various backgrounds is not helpful to train. \textcolor{redtarget}{In contrast,} the training can \textcolor{redtarget}{help when} their \textcolor{redtarget}{difficulty} is solved by \textcolor{redtarget}{augmentation (Figure 2(b), 2(c)).}[\ldots]}  \\\hline\hline
      \textbf{Rewriting\_medium} &\multicolumn{3}{|l}{\textbf{Modify the logical flow of ideas to improve the readability of the paragraph.}} \\
    \hline
      \multicolumn{2}{p{0.5\linewidth}|}{Patrick et al. \textcolor{bluesource}{proposed the Mouse Ether technique on finding} out that  when using multiple displays with different \textcolor{bluesource}{resolutions, a user loses the cursor because of} unnatural cursor movement between \textcolor{bluesource}{displays} [5]. The \textcolor{bluesource}{results showed that the} technique improved [\ldots]} & \multicolumn{2}{p{0.5\linewidth}}{Patrick et al. \textcolor{redtarget}{found} out that \textcolor{redtarget}{a user loses the cursor} when using multiple displays with different \textcolor{redtarget}{resolutions based on an} unnatural cursor movement between \textcolor{redtarget}{displays, and proposed a Mouse Ether technique} [5]. The \textcolor{redtarget}{proposed} technique improved [\ldots]}  \\\hline\hline
      \textbf{Rewriting\_heavy} &\multicolumn{3}{|p{0.75\linewidth}}{\textbf{Rewrite this paragraph to bring the argument through the idea that the goal is to learn a pixel-wise feature for semantic segmentation.}} \\
    \hline
      \multicolumn{2}{p{0.5\linewidth}|}{[\ldots] \textcolor{bluesource}{We consider propagating the labels from an annotated set} to \textcolor{bluesource}{an unlabeled set by nearest neighbor search in the featurespace. We} assume that  \textcolor{bluesource}{semantic  clustersemerge during training with sparse supervision, reinforced by aforementioned pixel-to-segment relationships. By propagating labels} in the feature space, we reinforce the \textcolor{bluesource}{learning of semantic clusters.}} & \multicolumn{2}{p{0.5\linewidth}}{[\ldots] \textcolor{redtarget}{Our goal is} to \textcolor{redtarget}{learn a pixel-wise feature that indicates semantic segmentation. It is thus reasonable to} assume that \textcolor{redtarget}{pixels and segments of the same semantics form a cluster} in the feature space, \textcolor{redtarget}{and} we reinforce \textcolor{redtarget}{such clusters with a featural smoothness prior: We find nearest neighbours in} the \textcolor{redtarget}{feature space and propagate labels accordingly.}}  \\\hline\hline
      \textbf{Concision and Rewriting\_light} &\multicolumn{3}{|p{0.75\linewidth}}{\textbf{Combine sentences 3 and 4 into a really short one keeping only the main idea. Improve the choice of wording.}} \\
    \hline
      \multicolumn{2}{p{0.5\linewidth}|}{[\ldots] Our method seeks to best approximate some  target \textcolor{bluesource}{distribution that is potentially multivariate,} using some chosen set of control \textcolor{bluesource}{distributions. We provide an implementation which gives unique, interpretable weights} in \textcolor{bluesource}{a setting of regular probability measures. For general probability measures, we construct our projection by first creating a regular tangent space through applying barycentric projection to} optimal \textcolor{bluesource}{transport plans.} Our application [\ldots] demonstrates the method’s \textcolor{bluesource}{efficiency and the necessity to have} a method that is applicable for general \textcolor{bluesource}{proabbility} measures. [\ldots]} & \multicolumn{2}{p{0.5\linewidth}}{[\ldots] Our method seeks to best approximate some \textcolor{redtarget}{general} target \textcolor{redtarget}{measure} using some chosen set of control \textcolor{redtarget}{measures. In particular, it provides a global (and} in \textcolor{redtarget}{most cases unique)} optimal \textcolor{redtarget}{solution.} Our application [\ldots] demonstrates the method’s \textcolor{redtarget}{utility in allowing for} a method that is applicable for general \textcolor{redtarget}{probability} measures. [\ldots]}  \\\hline\hline
      \textbf{Content\_deletion and Concision} &\multicolumn{3}{|l}{\textbf{Heavily remove details from this paragraph to make it more concise.}} \\
    \hline
      \multicolumn{2}{p{0.5\linewidth}|}{[\ldots] \textcolor{bluesource}{They should} only contain the name of the \textcolor{bluesource}{medication. Their design should be such that the user can decide whether} to \textcolor{bluesource}{add} or \textcolor{bluesource}{remove them from the display. [\ldots] On-calendar conflict representation should not be used as the main indication of an error after a rescheduling activity. The user} should \textcolor{bluesource}{instead be notified of the impending} conflict \textcolor{bluesource}{beforehand. Participants preferred that normal,} dismissible error messages \textcolor{bluesource}{be displayed and show} the \textcolor{bluesource}{full information regarding the conflicts being introduced by the action.} [\ldots]} & \multicolumn{2}{p{0.5\linewidth}}{[\ldots] These summaries should only contain the name of the \textcolor{redtarget}{medication and users should be able} to \textcolor{redtarget}{show} or \textcolor{redtarget}{hide them. [\ldots] The user} should \textcolor{redtarget}{be notified of a newly created} conflict \textcolor{redtarget}{upon rescheduling an entry, preferably via} dismissible error messages \textcolor{redtarget}{that describe} the \textcolor{redtarget}{conflict.} [\ldots]}  \\\hline\hline
  \end{tabular}
  \caption{Examples of revised paragraph with their associated annotation. Colouration based on difflib output.}
  \label{tab:examples}
\end{table*}
\section{Super-labels mapping}
\label{apx:supercat}
 In our taxonomy, boundaries between categories may be ambiguous, allowing for interpretation and discussion. Given this ambiguity, we defined super-labels that encompass categories of revision where similar actions are taken in Table~\ref{tab:mapping}. For example, the limit between \textit{Rewriting light} and \textit{Rewriting medium} or \textit{Content addition} and \textit{Development} can be blurry, and they totalise 59.43\% of complete disagreements (disagreement where there is no overlap between the two sets of labels). However, both opinions from annotators can be justified in discussions, as some paragraphs can be on the line of the two definitions. 

\begin{table*}
  \centering
  \begin{tabular}{lll}
    \hline
    \textbf{Super-label} & \textbf{Label} \\
    \hline
                   & Rewriting Light         \\
    Rewriting      & Rewriting Medium        \\
                   & Rewriting Heavy         \\\hline
    Concision  and   &  Concision  \\
    Content Deletion    &           Deletion\\\hline
      Development and             & Development      \\
      Content Addition & Content Addition  \\
                   &    Content Substitution  \\\hline
        Unusable  &   Unusable   \\\hline
  \end{tabular}
  \caption{Mapping between super-labels and labels}
  \label{tab:mapping}
\end{table*}

\section{Prompting}
\label{apx:prompt}
To work with the different models for revision, we use the following prompt (\textbf{\textcolor{blue}{Bold blue text}} correspond to the input data, the instruction and the paragraph to revise):\\

\texttt{You are a writing assistant specialised in academic writing. Your task is to revise the paragraph from a research paper draft that will be given according to the user's instructions.           
Please answer only by "Revised paragraph: <revised\_version\_of\_the\_paragraph>" \\
\textbf{\textcolor{blue}{instruction}} : \textbf{\textcolor{blue}{original\_paragraph}}}\\

For the comparative evaluation, based on the work of~\citep{raheja2023coedit}, the labels are mapped to general instructions, given in Table~\ref{tab:general_inst}.

\begin{table*}
  \centering
  \begin{tabular}{lll}
    \hline
    \textbf{Type} & & \textbf{Description} \\
    \hline
                   & Light         & Improve the English of this paragraph\\
    Rewriting      & Medium        & Rewrite some sentences to make them more clear and easily readable\\
                   & Heavy         & Rewrite and reorganize the paragraph for better readability\\\hline
    Concision      &               &Make this paragraph shorter\\\hline
       Content            & Deletion      & Remove unnecessary details\\\hline
  \end{tabular}
  \caption{Mapping of labels with general instructions}
  \label{tab:general_inst}
\end{table*}

\end{document}